%% file: SLMs-for-Curriculum-based-Guidance.tex
\documentclass[10pt]{article}

\usepackage[letterpaper]{geometry}
\usepackage{hicss}
\usepackage{times}
\usepackage[none]{hyphenat}
\usepackage{url}
\usepackage{latexsym}

\usepackage{indentfirst}
\usepackage{graphicx}
\graphicspath{{images/}}
\usepackage[round]{natbib}
\usepackage[table]{xcolor}
\usepackage{arydshln}
\usepackage[export]{adjustbox}

\usepackage[dvipsnames, svgnames]{xcolor}
\usepackage[T1]{fontenc}
\usepackage{soul}
\newcommand{\hlc}[2][yellow]{ {\sethlcolor{#1} \hl{#2}} }

\usepackage[breaklinks, hyperindex=true, colorlinks = true, urlcolor=blue, linkcolor=blue, citecolor=blue, bookmarks = true, backref=true]{hyperref}

\hypersetup{backref,
pdfpagemode=FullScreen,
colorlinks=true}

\usepackage{amssymb}

\title{Small Language Models for Curriculum-based Guidance }

 \author{Konstantinos Katharakis \\
  Copenhagen Business School \\ Frederiksberg, Denmark \\
  {\underline{\href{mailto:kok.digi@cbs.dk}{kok.digi@cbs.dk}} } \\ \And
  Sippo Rossi \\
  Hanken School of Economics \\  Helsinki, Finland \\
  {\underline{\href{mailto:sippo.rossi@hanken.fi}{sippo.rossi@hanken.fi}} } \\ \And
  Raghava Rao Mukkamala \\
  Copenhagen Business School \\ Frederiksberg, Denmark \\
  {\underline{\href{mailto:rrm.digi@cbs.dk}{rrm.digi@cbs.dk}} } \\ 
}

\date{}

\begin{document}
\maketitle
\begin{abstract}
\input{00-Abstract}

\end{abstract}

\subsubsection*{Keywords:}

Large Language Models, Small Language Models, Retrieval-Augmented Generation, Teaching Chatbots

\section{Introduction}
\input{01-Introduction}

\section{Related work} 
\label{sec:related-work}
\input{02-Related_work}

\section{Methodology}
\label{sec:methodlogy}

\input{03-Methodology}

\section{Results}
\label{sec:results}
\input{04-Results}

\section{Discussion}
\label{sec:discussion}

\input{05-discussion}

\section{Conclusion and future work}
\label{sec:conclusion}

\input{06-Future_work}
\bibliographystyle{plainnat}
\bibliography{sample}

\end{document}

%% file: 00-Abstract.tex
The adoption of generative AI and large language models (LLMs) in education is still emerging. In this study, we explore the development and evaluation of AI teaching assistants that provide curriculum-based guidance using a retrieval-augmented generation (RAG) pipeline applied to selected open-source small language models (SLMs). We benchmarked eight SLMs, including LLaMA 3.1, IBM Granite 3.3, and Gemma 3 (7–17B parameters), against GPT-4o. Our findings show that with proper prompting and targeted retrieval, SLMs can match LLMs in delivering accurate, pedagogically aligned responses. Importantly, SLMs offer significant sustainability benefits due to their lower computational and energy requirements, enabling real-time use on consumer-grade hardware without depending on cloud infrastructure. This makes them not only cost-effective and privacy-preserving but also environmentally responsible, positioning them as viable AI teaching assistants for educational institutions aiming to scale personalized learning in a sustainable and energy-efficient manner.

%% file: 01-Introduction.tex
Recent advancements in deep learning, particularly the emergence of Large Language Models (LLM), have generated great interest in exploring their potential applications beyond natural language processing~\citep{bommasani2021opportunities, rossi2024augmenting}. In similar lines, the advent of LLMs and generative AI tools has significantly impacted education by providing new opportunities for personalized learning tailored to each student's needs~\citep{deng2024does}. This shift toward personalized education not only enhances academic progress but also fosters a more engaging and interactive learning environment for students~
\citep{LLMs-CS-2023}. 
LLMs such as ChatGPT have rapidly transformed the way students seek explanations, examples, and feedback in academic contexts~\citep{KASNECI2023102274}. Their broad general-purpose knowledge and their ability to generate responses interactively make them attractive as on-demand tutors. 

However, several well-documented challenges hinder their effectiveness in students' pedagogical learning. First, LLMs often generate generic responses that may not adequately address the specific needs or contexts of individual learners. Not having customization can limit the effectiveness of personalized learning because, based on their level, students need customized explanations or examples that match their individual experiences and learning styles. Second, LLMs often produce fabricated or imprecise information, also known as hallucinations~\citep{huang2025survey}, which can mislead learners, adversely affecting their academic performance and overall learning experience~\citep{Shen_Wang_2024, Huirong_Xu_2024}.
Furthermore, closed LLMs usually do not offer the flexibility needed to align their outputs strictly with the course curriculum. This limitation presents a challenge for educators who want to keep students' focus on relevant course content. Lacking the capability to filter or direct the model's responses, learners may encounter irrelevant information, deviating from the intended learning objectives of the course. This not only distracts students from the purview of the course material but can also overwhelm them with superfluous details that are not relevant to the course.

Although closed source AI chatbots such as ChatGPT have shown potential as educational tools, they have significant limitations regarding personalized learning, factual reliability, and alignment with specific course curricula~\citep{cong2025systematic}. 
This lack of control results in most closed source AI chatbots and LLMs not providing enough control for educators. Therefore, there is a need to investigate how open-source small language models (SLM) could be augmented to bridge this gap by facilitating scalable, personalized, and course-aligned learning experiences. Therefore, the overarching research question for this study is:
\begin{quote}
\emph{How can open-source small language models be used to build an AI teaching assistant that provides students with personalized learning and curriculum-based guidance?}
\end{quote}

To address the above research question, we explore the development and evaluation of generative AI teaching assistants using open-source SLMs that are designed to enhance student learning through curriculum-based guidance. We developed and tested eight AI teaching assistants based on open-source SLMs that utilize a retrieval-augmented generation (RAG) pipeline, which indexes the official course curriculum, such as teaching materials and lecture slides from a graduate level mathematics, statistics and linear algebra course lectured in a Scandinavian university. During inference, the system identifies and retrieves the most relevant segments and pointers from course readings and slides to answer a student's query. These chunks of text are then integrated into the model's response, complete with citations, to maintain transparency and improve factual grounding.

The design using SLMs offers several advantages over commercial and closed source models such as GPT-4o. First, curriculum consistency is achieved as responses are tightly coupled to approved materials by the educators, ensuring alignment with the course's reading materials and learning outcomes. Second, the prevalence of erroneous or imprecise information will be minimized, as the responses are guided towards being based on verifiable teaching materials, thereby enhancing factual accuracy and reducing speculative or hallucinated output. 
Third, the performance is better in terms of cost and required computational resources, as each model operates with less than 17 billion parameters, allowing for deployment on consumer-grade GPUs or institution-hosted servers. 
Lastly, the use of open-source architecture, with public model weights and evaluation scripts facilitates consistent performance, easy customization and future development.

Our study offers several contributions. First, it showcases the practical implementation of AI teaching assistants powered by open-source SLMs, designed for personalized and curriculum-based guidance in higher education. Second, we demonstrate the usefulness of the RAG pipeline, which improves factual accuracy and alignment with official teaching materials by reducing hallucinations, enhancing both accuracy and pedagogical relevance. Lastly, we present a scalable and privacy-conscious alternative to commercial LLMs by using lightweight models (with $\leq$17B parameters) suitable for real-time inference on local institutional servers, in an infrastructure typically found in medium to large universities. 

The rest of the paper is organized as follows. In section two, we present the related work, followed by a description of the methodology in section three. In section four, we present the results of our experiments with the SLMs, and in section five we discuss these results. In section six, we conclude the paper with remarks on future work.



%% file: 02-Related_work.tex
\begin{figure*}[h!]
    \centering
    \includegraphics[width=0.9\textwidth]{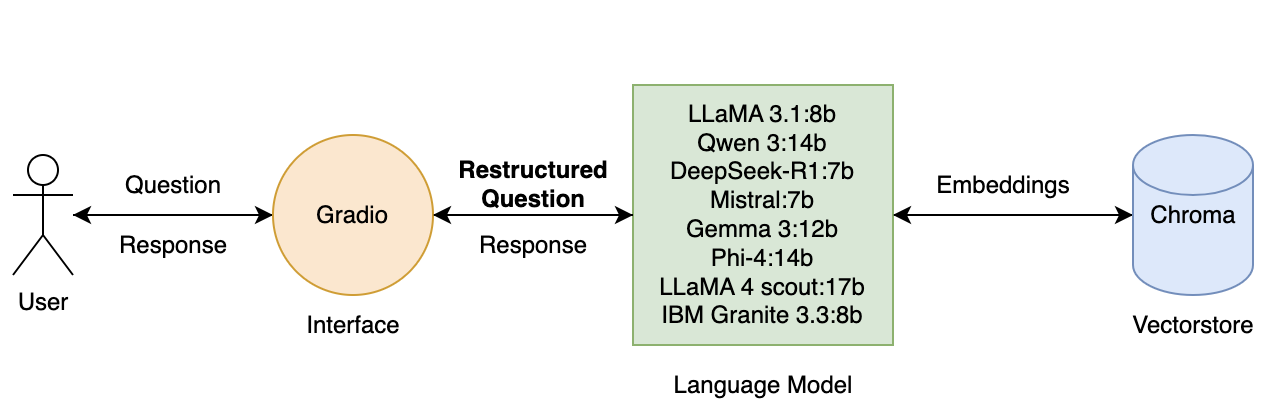}
    \caption{Final architecture}
    \label{fig:final_architecture}
\end{figure*}

Recent studies on educational AI have demonstrated the pedagogical benefits of conversational AI teaching assistants powered by language models and retrieval-augmented generation (RAG). In this section, we outline literature that influenced the methodological choices of this paper. It should be noted that due to the recency of AI teaching assistants, peer-reviewed literature is still scarce, and many publications are still pre-prints or part of conference proceedings. 

As the primary inspiration for this paper, Harvard's CS50 course deployed a suite of LLM-driven tools, including a chatbot called CS50 Duck \citep{10.1145/3626253.3635427} that provides code explanations and debugging support while preserving academic integrity. Course participants described it as akin to having a "personal tutor", and the results indicate that AI integration in education enhances the learning experience. Building on their earlier work, with the purpose of improving the AI tools used in CS50, \cite{liu2025improving} developed a human-in-the-loop framework for continuous improvement of LLM-based systems. A key finding of the study was that fine-tuning with small, high-quality datasets significantly improved alignment of the models.
%
%
Earlier papers have also investigated the use of AI-based TAs in introductory programming education \citep{lee2023learningteachingassistantsprogram}. \cite{lee2023learningteachingassistantsprogram} found that learners completed tasks more quickly while achieving similar performance when assisted by AI teaching assistants. Learners rated the AI TAs on par with human TAs in terms of response speed, comprehensiveness, and overall satisfaction. 

However, there have also been studies that have shown less positive results. \cite{tsai2023exploring} explored the integration of ChatGPT into chemical engineering education to enhance problem solving and critical thinking. The authors propose an LLM-assisted approach to deepen understanding and support skill development. An experimental lecture and student projects revealed that while LLMs are accessible and practical, student feedback was mixed. The authors advocate for active adoption of LLMs in curricula, emphasizing the need for critical thinking and responsible use. As another example, an analysis on LLM use in an advanced computing class in India \citep{arora2025analyzing} raised concerns about over-reliance on LLMs for tasks such as code generation, debugging, conceptual understanding, and test case creation, particularly when students submitted entire assignments as prompts in attempts to receive full solutions. 



Furthermore, \cite{Huirong_Xu_2024} conducted a systematic review of LLMs for education, noting that LLMs are ill-suited for mathematical tasks, being limited to replicating patterns learned from training data. To mitigate these limitations, they propose fine-tuning on synthetic arithmetic tasks, enhancing logical reasoning, and integrating external tools such as language converters, information retrieval systems, and calculation engines. They outline two development pathways: advancing foundational research on LLMs' mathematical capabilities and promoting inclusive mathematics education through personalized learning experiences.
Similarly, \cite{Shen_Wang_2024} address the challenges of implementing LLMs like GPT-4o in real-world educational settings despite their promising experimental performance. They propose using RAG techniques to provide LLMs with necessary prior information and guide them toward pedagogically appropriate outputs. 
\cite{article} examine how Generative AI tools are reshaping education in information systems and business analytics undergraduate education, calling for educators to update their teaching methods to teach students to critically assess GAI outputs, not just use them. 

When compared to the existing literature, our work adds to the growing field of AI educational tools by showing that open-source small language models, when used with a retrieval-augmented generation system, can perform as well as or even better than large proprietary language models like GPT-4o in providing personalized student support aligned with the course curriculum. While previous studies, such as Harvard’s CS50 Duck~\citep{10.1145/3626253.3635427}, have looked at AI teaching assistants using large models, our method uses lightweight, open models ($\le$17B parameters) to develop scalable, real-time teaching assistants that can be easily deployed on standard hardware within the university premises. 
Moreover, we emphasize strict adherence to the curriculum and uphold academic integrity by limiting model responses to course content. This combination of open-source utilization, cost-effectiveness, and pedagogical emphasis distinguishes our study as we examine the practical implementation of AI in higher education.




%% file: 03-Methodology.tex
As shown in Figure~\ref{fig:final_architecture}, we developed a total of eight AI teaching assistants to evaluate the performance of different open-source SLMs and their suitability for use in a mathematics course. We also tested one closed-source proprietary LLM. The goal of this setup was to compare the effectiveness of SLMs against a state-of-the-art LLM in the same task. 
The open-source SLMs used included LLaMA 3.1 \citep{grattafiori2024llama3herdmodels} with 8 billion parameters, Qwen 3 \citep{yang2025qwen3technicalreport} with 14 billion, DeepSeek-R1  \citep{deepseekai2025deepseekr1incentivizingreasoningcapability} with 7 billion, Mistral \citep{jiang2023mistral7b} with 7 billion, Gemma 3 \citep{gemmateam2025gemma3technicalreport} with 12 billion, Phi-4 \citep{abdin2024phi4technicalreport} with 14 billion, LLaMA 4 Scout \footnote{\url{https://ai.meta.com/blog/llama-4-multimodal-intelligence/}} with 17 billion and IBM Granite-3.3 \footnote{\url{https://huggingface.co/ibm-granite/granite-3.3-8b-instruct}} with 8 billion parameters. IBM Granite-3.3 was downloaded through Hugging Face \footnote{\url{https://huggingface.co}} while the rest of the 7 SLMs were downloaded through Ollama \footnote{\url{https://ollama.com}}. The model parameters were left at default, except for the temperature, which was set to 0.4 for all models. The parameters are provided in Table 1. 

\begin{table*}
\begin{center}
\label{table:model-parameters}
\caption{Model parameters}
\begin{tabular}{|l|c|c|c|}
\hline
 \textbf{Model} & \textbf{Top-P} & \textbf{Top-K} & \textbf{Temp} \\ 
 \hline

 LLaMA 3.1 & 1 & 40 & 0.4 \\ 
 \hline

 Qwen 3 & 0.8 & 20 & 0.4 \\ 
 \hline

 DeepSeek-R1 & 0.95 & Not set & 0.4 \\ 
 \hline

 Mistral & 0.9 & Not set  & 0.4 \\ 
 \hline

 Gemma 3 & 0.6 & 50 & 0.4 \\ 
 \hline

 Phi-4 & 0.95 & 40 & 0.4 \\  
 \hline

 LLaMA 4 Scout & 0.6 & 0.4 & 0.4  \\  
 \hline
 
 IBM Granite-3.3 & Not set & Not set & 0.4  \\  
 \hline

 GPT-4o & 1 & Not set & 0.4  \\  
 \hline
 
\end{tabular}
\end{center}
\end{table*}

These models were selected to represent a diverse range of top-performing open models, with an emphasis on using the smallest viable version from each model family that could answer questions related to the course contents with satisfactory performance. The variety allowed us to assess which SLMs offered the best trade-off between performance and computational efficiency when deployed as part of a RAG framework supporting course-specific queries. The LLM used for benchmarking was OpenAI's $\sim$200B parameter GPT-4o \citep{openai2024gpt4ocard}. GPT-4o's role was to serve as a reference point/baseline, allowing us to evaluate the relative performance of each SLM. By comparing their output to those of GPT-4o, we could assess how closely each SLM approximated the capabilities of a leading proprietary model.
\subsection{Data}
The corpus used in the RAG-pipeline was derived from lecture slides, comprising a total of 726 slides and other reading materials. The language of all the content on the slides was English. To ensure the relevance and clarity of content, the slides were pre-processed first by turning them from .pdf to a .txt form, then by removing any general information or visuals that were not directly related to the specific lecture material. For slides containing images, each image was passed through GPT-4o mini \citep{openai2024gpt4ocard} to generate detailed textual descriptions of the depicted content. These descriptions were then automatically inserted into the slide text at the original image positions using a custom pipeline. This approach preserved the narrative flow of the lecture content while making visual information accessible in text form for the language models. The final vector store was created using the chroma module\footnote{\url{https://www.trychroma.com/}} with the OpenAI embeddings model\footnote{\url{https://python.langchain.com/docs/integrations/text_embedding/openai/}}.

\subsection{Architecture}
The system message used in this study was inspired by the CS50 duck system message \citep{10.1145/3626253.3635427}. The system message used was the following: 
\begin{quote}
\hlc[Beige]{
\textit{You are a friendly and supportive teaching assistant for [name of the course] at [name of institution]. You answer student questions about linear algebra, statistics and data science. You also answer to questions about the course administration of the class. Do not answer questions about unrelated topics. You provide guidance through brief, concise answers with clear steps. You must not give direct answers, as this upholds academic honesty. Your goal is to encourage the student to think critically.}}
\end{quote}
Each of the AI teaching assistants was created using helper classes and functions from LangChain \citep{Chase_LangChain_2022}. LangChain is an open-source framework designed to simplify the development of LLM applications, enabling automated workflows by providing tools for chaining together components like prompts, models, and data sources. The goal was to start with a simple architecture and improve it until it reached a threshold where at least 50\% of the models would respond to over 50\% of the questions in a way that aligned with the system prompt, the conversation context, and the materials retrieved from the vector store.
%
%
In the initial architecture of the model, when the user submits a question through the gradio interface \citep{Abid_Gradio_Hassle-free_sharing_2019}, the system then embeds the query and retrieves the most semantically similar documents from the course corpus using vector similarity. These relevant documents are then provided as context to the language model, which generates a response based on both the user's question and the retrieved materials.

However, we noticed that the models were not consistently producing responses aligned to the goal of not providing a direct solution but instead instructions on how to solve task. For this reason, as a new step, a fixed text was added before passing the question of the user, which reminded the model that it should not give a solution to the user, and that its purpose is to guide the student using thorough and clear steps. In other words, each prompt was adjusted to contain the following: \textit{Don't solve this for me. Instead, guide me through the process by outlining the steps I should follow to reason through it on my own: $<$Original Prompt$>$}. Figure~\ref{fig:final_architecture} shows the architecture of the AI teaching assistant.



\begin{table*}[htbp]
\centering
\caption{Model performance on three categories of validation questions (\(\# \)  of correct responses out of 6)}
\smallskip
\begin{tabular}{|l|c|c|c|}
\hline
\textbf{Model} & \textbf{Conversation} & \textbf{RAG Pipeline} & \textbf{System Message}\\
 & \textbf{Tracking} & \textbf{Usage} & \textbf{Following} \\\hline
DeepSeek-R1:7b & 3 & 6 & 1 \\
\hline
Mistral:7B & 6 & 6 & 5 \\
\hline
IBM Granite 3.3:8b & 6 & 4 & 4 \\
\hline
LLaMA 3.1:8b & 2 & 0 & 3 \\
\hline
Gemma 3:12b & 6 & 2 & 3 \\
\hline
Qwen 3:14b & 6 & 6 & 5 \\
\hline
Phi-4:14b & 5 & 6 & 3 \\
\hline
LLaMA 4 scout:17b & 2 & 6 & 0 \\
\hline
GPT-4o & 6 & 6 & 5 \\
\hline
\end{tabular}

\label{tab:model_performance}
\end{table*}

\subsection{Validation during development}

Before proceeding to the main assessment of the models and the AI teaching assistant, we first did an initial validation of the response quality for each of the models to guide development of the architecture. The response quality was assessed with a set questions to evaluate retrieval accuracy (i.e., whether the system correctly uses the vector database), to assess whether the model adheres to the system message, and to evaluate memory (consisting of a primary question followed by a related follow-up) to check if the model retains context. We then tested all possible orders and combinations of these questions, keeping the memory pair together to preserve their contextual link, resulting in 6 (3!) permutations. In each iteration, we recorded whether the model gave the correct response to each question. Table~\ref{tab:model_performance} shows, out of the 6 total runs, how many times each model produced a response aligned to the previous conversation, the database or the system prompt. The initial assessment showed that the models did not provide consistent responses across different runs and that in some cases their accuracy depended on the order in which the questions were asked. GPT-4o was used, for early benchmarking purposes, to compare the SLMs with a state-of-the-art LLM. 

\begin{figure*}[t]
    \centering
    \includegraphics[width=0.7\textwidth]{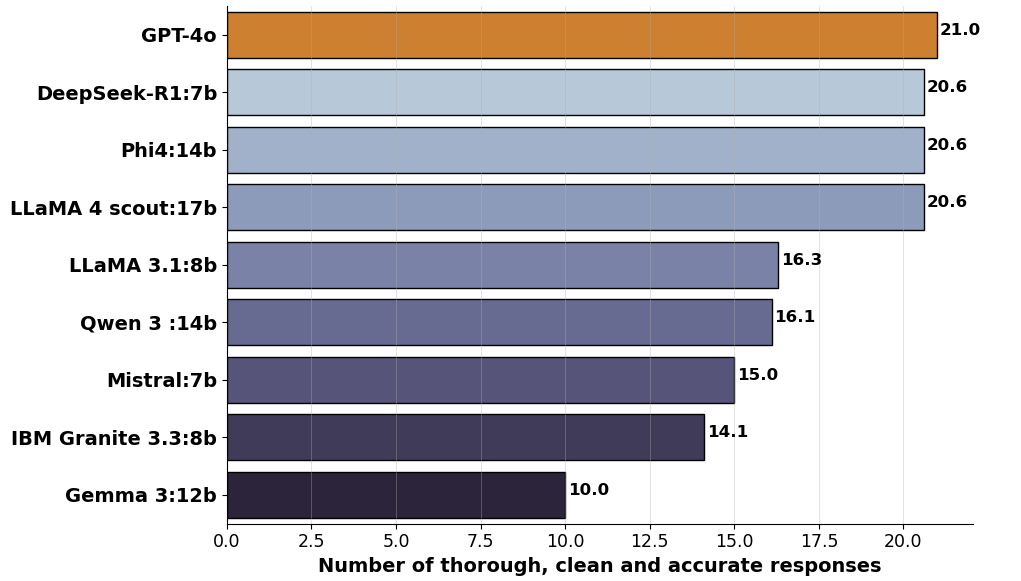}
    \caption{Average performance of models on theory questions based on running the model 10x}
    \label{fig:theory_questions_performance}
\end{figure*}

%
%

For the conversation context maintenance assessment, or in other words checking if the model remembered the previous context of the conversation, we made a set consisting of a main question and a follow-up question, where first question was “What is an eigen value?” followed by the question “What is its relation to machine learning?”. To evaluate performance, we manually verified and counted how many times each model correctly addressed the relationship between eigenvalues and machine learning, rather than for example giving a generic answer about machine learning or linear algebra.  The Conversation Tracking column in Table~\ref{tab:model_performance} shows the results of the conversation context maintenance assessment. In a test of six conversational turns, four out of the eight models (IBM Granite 3.3, Qwen 3, Mistral and Gemma 3) successfully remembered the previous message in all six cases. Phi-4 and DeepSeek-R1 showed slightly lower consistency, with Phi-4 succeeding in 5 out of 6 cases and DeepSeek-R1 in 3 out of 6. LLaMA 3.1 and LLaMA 4 succeeded in 2 out of the 6 cases. GPT-4o, as expected, successfully remembered the past context in all six cases. These findings suggested that the initial architecture showed potential for building a conversational AI teaching assistant that handles follow-up questions maintaining continuity in multi-turn interactions when guiding students through the topics in the course curriculum.

For the retrieval assessment, we used the question: "Who is $<$the name of instructor of the class$>$?" where the name of the instructor of the class is a placeholder referring to the course instructor, whose details were included in the lecture slides. The goal was to determine whether the models could accurately extract and return information such as the instructor's full name, email address, and other relevant details from the course materials. We evaluated each model by counting how many times it correctly retrieved and presented the information.  In this test, we focused on evaluating how well the SLMs cooperated with the embedding model in our architecture. Specifically, we assessed whether the models could retrieve and use relevant information from the course material in response to the user queries over six conversational turns. As depicted in the RAG Pipeline Usage column in Table~\ref{tab:model_performance}, five out of the eight models (Qwen 3, DeepSeek-R1, Mistral, Phi-4, and LLaMA 4) successfully retrieved the correct context in all six cases. Gemma 3 showed weaker performance, responding according to the retrieved relevant information only in 2 out of 6 cases, while LLaMA 3.1 failed to give a response aligned to the retrieved information in all cases. GPT-4o, again as expected performed accurately in all cases. These results demonstrate that student queries are addressed accurately and contextually according to the course curriculum.

Finally, we evaluated how well each model adhered to the system prompt by asking the models to assist with finding the null space of a given matrix (the model instructions can be found on the methodology section). For this assessment, we counted how many times the models followed the system prompt, offering guidance rather than providing the full solution. As can be seen in the System Message Following column in Table~\ref{tab:model_performance}, Qwen 3 and Mistral followed the prompt in 5 out of 6 cases, while Granite IBM did so in 4 out of 6. LLaMA 3.1, Gemma 3, Phi-4 offered guidance in 3 out of 6 cases while DeepSeek-R1 in one case and LLaMA 4 in none of the 6. GPT-4o failed once to provide guidance instead of a solution. Based on these results, we concluded final architecture was functional and capable of supporting a conversational AI assistant.

\begin{table*}[h]
\centering
\caption{Course assignment questions in linear algebra and statistics}

\begin{tabular}{|c|p{15cm}|}
\hline

\multicolumn{2}{|c|}{\textbf{Theoretical Questions}} \\ \hline

\textbf{No.} & \textbf{Question} \\
\hline
1 & What conditions ensure that a system of linear equations has a unique solution? \\
\hline
2 & How does the span of a set of vectors relate to the concept of feature space in machine learning models? \\
\hline
3 & Why is linear independence crucial for forming a basis of a vector space? \\
\hline
4 & In what ways does the rank–nullity theorem help explain the structure of linear mappings? \\
\hline

\multicolumn{2}{|c|}{\textbf{Course Assignment Questions}} \\ \hline

\textbf{No.} & \textbf{Question} \\
\hline

1 & a. Define $g:\mathbb{Z} \to \mathbb{Z}$ by the rule $g(n)=4n-5$, for all integers $n$. \\
 &    (i) Is $g$ one-to-one? Prove or give a counterexample. \\
 &    (ii) Is $g$ onto? Prove or give a counterexample. \\
 & b. Define $G:\mathbb{R} \to \mathbb{R}$ by the rule $G(x)=4x-5$ for all real numbers $x$. Is $G$ onto? Prove or give a counterexample. \\
\hline
2 & Please answer the following questions with proper explanation. \\
 & 1. Suppose a $5 \times 6$ matrix $A$ has four pivot columns. What is dim Nul $A$? Is Col $A = \mathbb{R}^4$? Why or why not? Please discuss. \\
 & 2. If $A$ is a $4 \times 3$ matrix, what is the largest possible dimension of the row space of $A$? If $A$ is a $3 \times 4$ matrix, what is the largest possible dimension of the row space of $A$? Explain. \\
 & 3. If $A$ is a $6 \times 4$ matrix, what is the smallest possible dimension of Nul $A$? Please discuss your solution. \\
 & 4. If $A$ is a $6 \times 3$ matrix and if we know that its rank is 3 then what is the dimensions of Nul $A$, dimensions of Row $A$ and dimensions of Col $A$? Please discuss your solution. \\
\hline
3 & A factory produces widgets. 5\% of these widgets are defective. A test can detect defective widgets with a 95\% accuracy rate (i.e., it correctly identifies defective widgets 95\% of the time). However, the test also has a 2\% false-positive rate, meaning it incorrectly classifies a non-defective widget as defective 2\% of the time. If a widget tests positive, what is the probability that it is actually defective? \\
\hline
\end{tabular}
\label{tab:assignment-questions}
\end{table*}

%% file: 04-Results.tex

After validating the architecture's suitability, we conducted a comprehensive evaluation of each AI teaching assistant using a battery of questions. The assessment used 29 assignment-derived questions and 21 broader theoretical questions specifically crafted to align with the course's learning objectives and which could be asked in the course's oral exam. The theoretical questions were designed to assess the models' ability to provide contextually accurate and course-relevant responses. Examples of the theoretical questions and course assignment questions are provided in Table~\ref{tab:assignment-questions}. The course covered topics in both statistics and linear algebra, and the evaluation set contained 17 questions related to statistics and 33 questions related to linear algebra, where the uneven distribution is due to the course having more lectures and tasks related to the latter subject. Each model was queried 10 times per question set and results are reported as averages.

The authors manually evaluated the responses to both the course questions and theoretical questions. We also calculated a hallucination rate for the models, with the average hallucination rate being 37.19\% across all models prior to introducing the RAG-pipeline, while when using a RAG-pipeline the hallucination rate dropped to 0\%. Moreover, the SLMs' responses to all questions were compared to the answers produced by GPT-4o. The purpose of this method is to benchmark the number of valid SLM responses and compare them against those of a state-of-the-art model.

Figure~\ref{fig:theory_questions_performance} presents the average results of the responses of the SLMs in the theory questions. LLaMA 4, Phi-4, and DeepSeek-R1 achieved an average of 20.6 theoretical questions accurately and thoroughly, providing concise responses that did not mislead the student. LLaMA 3.1 and Qwen 3 followed closely, with averages of 16.3 and 16.1 accurate and comprehensive responses out of 21. On average, Mistral, Granite 3.3, and Gemma3 performed well in 15, 14.1, and 10 cases respectively. 
Overall, three SLMs, LLaMA 4, Phi-4, and DeepSeek-R1 matched GPT-4o's performance, demonstrating that properly configured SLMs were capable of guiding students effectively through the theoretical components of the course curriculum. 
\begin{figure*}[t]
    \centering
    \includegraphics[width=0.7\textwidth]{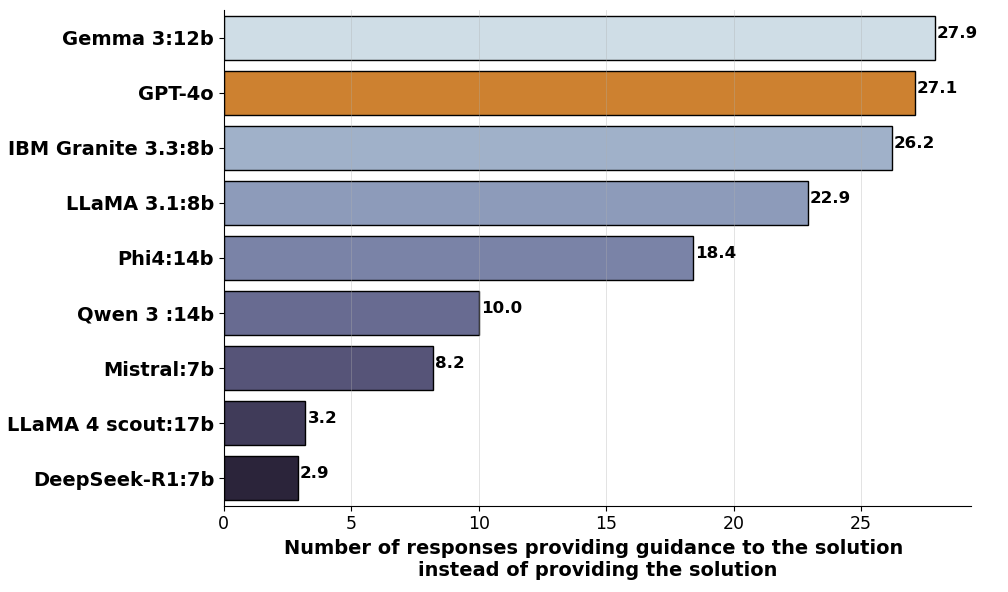}
    \caption{Average performance of models on course assignment/guidance questions based on running the model 10x}
    \label{fig:assignment_questions_performance}
\end{figure*}

The average results of the models when answering course assignment questions are shown in Figure~\ref{fig:assignment_questions_performance}. Gemma 3 was the top-performing model, surpassing even GPT-4o. It averaged 27.9 successful responses out of 29 cases, offering clear step-by-step guidance while avoiding direct solutions. Granite 3.3 also performed well, averaging 26.2 correct responses, just 0.9 fewer than GPT-4o which scored 27.1 out of 29. 
LLaMA 3.1 averaged 22.9 appropriate responses, while Phi-4 responded correctly in 18.4 out of 29 evaluated cases. Performance dropped significantly with the remaining models: Qwen 3 succeeded in 10 instances, Mistral in 8.2, and LLaMA 4 in only 3.2. DeepSeek-R1 responded correctly in 2.9 out of the 29 cases. 
These results emphasize that with a simple prompt-based adjustment, SLMs such as Gemma 3 and Granite 3.3 were able deliver high-quality educational guidance suitable for use as a course assistant.

Overall, the averaged results demonstrate that open-source SLMs can serve as teaching assistants when configured with a well-crafted system prompt, which can reliably steer model behaviour in line with pedagogical goals. 

%% file: 05-discussion.tex
The findings of our study highlight the potential of open-source SLMs in providing high-quality responses aligned with curricula, matching the capabilities of advanced closed-source LLMs like GPT-4o in specific scenarios. Specifically, in our limited test set, the Gemma 3 model (Figure~\ref{fig:assignment_questions_performance}) performed best in handling course assignment questions, indicating that SLMs can deliver accurate and pedagogically suitable guidance, provided they are configured appropriately. These findings are particularly notable since these models operated with a fraction of the computational power and a fraction of the parameters needed by GPT-4o, emphasizing their practical benefits for real-time applications in typical university settings without dependence on commercial cloud APIs.

The differences in performance among models also highlight important design considerations. For instance, while DeepSeek-R1. Phi-4,  and LLaMA 4 excelled in theoretical inquiries (Figure~\ref{fig:theory_questions_performance}), they faced considerable challenges with assignment-related prompts, suggesting potential shortcomings in instructional alignment or task generalization. In contrast, Gemma 3 performed well on assignments but did not do as well in retrieval tasks during earlier assessments. These variations indicate that a model's architecture and training data impact not only its overall performance but also the types of educational interactions it is most effective in, such as conceptual explanation, curriculum retrieval, and guidance.

Our results also suggest the importance of prompt engineering and context design in educational AI systems. Adding explicit system messages and structured guidance within prompts effectively directed model behavior away from providing direct answers and instead toward curriculum-based guidance. 
This approach of providing guidance instead of direct solutions will help support academic integrity and align with pedagogical best practices. The RAG pipeline attempted to ensure the model's outputs were based on reliable course material, which reduced errors and aligned with educational goals. This is crucial in formal educational practices, where accuracy and context are vital.
Finally, using open-source SLMs helps ensure transparency and reproducibility, providing local control over how they are used, which is crucial for educational institutions that prioritize data privacy, cost-effectiveness, and the freedom to choose teaching methods. By evaluating 8 SLMs and demonstrating that some can perform approximately at the level of GPT-4o in certain situations, our study encourages schools to create affordable and customizable AI teaching assistants. These systems can be adapted for specific courses while fitting within existing infrastructure, providing a practical solution for improving education with AI.

One of the primary sustainability benefits of SLMs compared to LLMs is their significantly lower energy consumption during both training and inference. The models in our study, with 7 to 17 billion parameters, ran effectively on consumer-grade GPUs or institution-owned servers. This is very different from LLMs like GPT-4, which need large, energy-intensive cloud platforms. This lower computing needs leads to reduced carbon emissions and energy costs, making SLMs a more eco-friendly option for educational institutions, especially when used across multiple courses or departments. Additionally, local deployment avoids the energy drain caused by frequent API requests to remote servers, helping create greener, decentralized AI systems for education. In this way, SLMs not only make AI teaching assistants more accessible but also support institutional goals of sustainability and responsible computing.

Due to the proof-of-concept nature of the AI teaching assistant presented in this paper, the emphasis was on developing a working architecture and identifying suitable models with limited testing. The results presented in this paper demonstrate the usability of SLMs in a mathematics course, and thus, the results could differ for courses in other disciplines. Consequently, the generalizability of the results is limited, which we plan to address in future work.

Moreover, this study encountered several limitations that affected the effectiveness and generalizability of AI teaching assistants. First, the use of SLMs inherently restricted the system's capabilities because of their limited number of parameters and smaller context windows. These limitations hindered the models' ability to reason over longer passages. Second, resource constraints prevented us from testing and deploying larger, more powerful open-source models. The computational and memory requirements of models exceeding 17B parameters exceeded the available infrastructure, necessitating a focus on lightweight models, even if they involved trade-offs in accuracy or depth of response. Finally, the subject domain of linear algebra presents intrinsic challenges for AI-based tutoring. Its abstract concepts and the need for multi-step reasoning, along with a strong reliance on visual and symbolic representations, make it a difficult topic for LLMs to manage, especially in the limited context of open-source SLMs.

%% file: 06-Future_work.tex
The results of our study indicate that with proper alignment and architecture, SLMs can deliver performance appropriate for use as course teaching assistants in low-resource environments like universities and schools, while maintaining energy efficiency and a low carbon footprint.  
Our tests using 50 multi-turn theory and course assignment tasks grounded by a RAG pipeline show that the eight open-source models with 7 - 17 billion parameters could achieve results similar to GPT-4o, while running in real time on a single GPU server. However, due to the limited nature of the tests and evaluation presented in this paper, further validation is needed through broader future studies.
Nevertheless, these findings highlight the potential for further research on locally hosted, resource-efficient AI tutors using open-source SLMs.

Building on the results of this study, our next steps will focus on refining and scaling the AI teaching assistant system. We plan to fine-tune the top-performing SLMs using domain-specific data to improve their pedagogical alignment and contextual reasoning capabilities. Furthermore, we aim to experiment with larger variants of the LLaMA, Qwen, and Gemma to evaluate their suitability for deployment. These larger models, with expanded parameter counts and context windows \citep{yang2025qwen3technicalreport, grattafiori2024llama3herdmodels,gemmateam2025gemma3technicalreport}, are expected to handle complex queries more robustly and support multi-turn conversations with improved accuracy and retention. 

In addition to developing the models, we plan to pilot the AI teaching assistant in multiple courses during the autumn of 2025, where we will conduct a structured evaluation using surveys, interviews, and by monitoring learning performance metrics. Insights from this pilot will inform iterative improvements before broader deployment across other departments at the university, and provide data for more robust future studies.